\documentclass{article}

\usepackage{graphicx}
\usepackage{algorithm}

\PassOptionsToPackage{numbers,sort&compress}{natbib}
\usepackage[preprint]{neurips_2025}

\usepackage{algpseudocode}
\usepackage[endLComment=]{algpseudocodex}
\usepackage{amsmath}
\usepackage{multirow}

\usepackage[utf8]{inputenc} 
\usepackage[T1]{fontenc}    
\usepackage{hyperref}       
\usepackage{url}            
\usepackage{booktabs}       
\usepackage{amsfonts}       
\usepackage{algorithm}      
\usepackage{algpseudocode}  
\usepackage{nicefrac}       
\usepackage{enumitem}
\usepackage{xcolor}
\usepackage{wrapfig}

\usepackage{minted}

\usepackage{microtype}      

\definecolor{darkgreen}{RGB}{0,100,0}
\colorlet{sectioncolor}{darkgreen}

\title{ProfilingAgent: Profiling-Guided Agentic Reasoning for
Adaptive Model Optimization
}
\author{
  Sadegh Jafari \quad Aishwarya Sarkar \quad Mohiuddin Bilwal \quad Ali Jannesari \\
  Department of Computer Science \\
  Iowa State University \\
  \texttt{\{sadegh, asarkar1, moh596, jannesar\}@iastate.edu}
}

\begin{document}

\maketitle

\begin{abstract}
Foundation models face growing compute and memory bottlenecks, hindering deployment on resource-limited platforms without optimization. While model compression techniques like pruning and quantization are widely used, most existing approaches rely on uniform heuristics or rule-based strategies that ignore architectural and runtime heterogeneity. Profiling tools have been developed to expose such bottlenecks by capturing metrics like per-layer latency, memory usage, and compute cost; however, these insights are rarely integrated into automated compression pipelines. In this paper, we propose ProfilingAgent, a profiling-guided agentic approach that leverages large language models (LLMs) to automate model compression through structured pruning and post-training dynamic quantization. Our modular pipeline consists of a multi-agent system that leverages both static (e.g., MACs, parameter counts) and dynamic (e.g., latency, memory usage) profiling signals to generate architecture-specific compression strategies. Unlike heuristic baselines for pruning and quantization, our LLM-guided agents reason over profiling traces to produce layer-wise decisions tailored to performance bottlenecks. Experiments conducted on benchmark datasets, such as ImageNet 1K, CIFAR-10 and CIFAR-100, utilizing ResNet-101, ViT-B/16, Swin-B, and DeiT-B/16, demonstrate that our approach to pruning maintains competitive or even improved accuracy (with accuracy drops of about 1\% on ImageNet-1K, and gains of up to +2\% for ViT-B/16 on smaller datasets without any post-pruning fine-tuning), while quantization achieves memory savings of up to  74\% with accuracy drops below 0.5\%. Furthermore, our quantization strategy consistently delivers inference speedups of up to 1.74×, effectively reducing latency while preserving performance. Comparative studies with GPT-4o and GPT-4-Turbo underscore the impact of LLM reasoning quality on iterative pruning efficacy. These results establish agentic systems as a scalable solution for profiling-guided model optimization.

\end{abstract}

\section{Introduction}
Modern deep neural networks have achieved remarkable progress in vision, language, and multimodal tasks, but their growing scale imposes substantial computational and memory costs \cite{li2023model}, limiting deployment on latency or resource-constrained platforms. To address this, compression techniques such as pruning, quantization, low-rank approximation, and distillation have become essential for reducing model size, memory footprint, and inference latency while preserving accuracy \cite{msuya2023deep}. Among these, pruning \cite{kuzmin2023pruning} and quantization \cite{fan2020training} are most prevalent due to their hardware compatibility and ease of deployment. Pruning removes redundant neurons, filters, or blocks to produce sparse networks \cite{gholami2022survey}. Unstructured pruning offers fine-grained sparsity but yields memory-bound sparse matrix operations that are difficult to accelerate \cite{buluc2008challenges, gale2019state}. Structured pruning, in contrast, removes entire channels or heads, maintaining dense computation and offering practical speedups with minimal accuracy loss. Quantization reduces the bit-width of weights and activations for smaller memory usage and faster inference. While quantization-aware training (QAT) accounts for precision loss during training, post-training quantization (PTQ) applies after training and risks greater accuracy degradation, particularly when uniformly applied across heterogeneous layers \cite{zhao2023post}.

Despite their promise, pruning and quantization are often applied without considering a model’s runtime profile. Heuristic or uniform compression can degrade accuracy or yield minimal gains, particularly in architectures with heterogeneous layer behavior or hardware bottlenecks. This highlights the need for profiling-aware pipelines that adapt compression strategies to actual layer-wise performance. Tools like PyTorch Profiler \cite{pytorch_profiler}, TensorFlow Profiler \cite{tensorflow_profiler}, and NVIDIA DLProf \cite{nvidia_profiler} provide detailed layer-level metrics, including FLOPs, latency, memory usage, and parameter counts, enabling identification of compute or memory-intensive bottlenecks \cite{gyawali2023comparative, nvidia_dlprof_pyprof}. However, these insights are rarely integrated into compression workflows, where static, black-box strategies remain the norm. Moreover, manually tuning compression schemes for diverse architectures, ranging from ResNet to hybrid transformers like Swin and DeiT, is labor-intensive and brittle, often requiring expert intuition and extensive trial-and-error to balance trade-offs
between accuracy, latency, and memory. Each model exhibits unique bottlenecks and inter-layer dynamics, making manual optimization inefficient and error-prone. To address this, we propose the use of LLM-guided AI agents that autonomously interpret profiling data and generate targeted pruning and quantization strategies. These agents reason over architectural structure and runtime behavior to outperform static heuristics and avoid expensive grid searches. Unlike rule-based approaches, agentic systems operate autonomously, adapt to context, and iteratively refine their decisions to optimize performance.
Agentic approaches, particularly those built on LLMs, represent a significant shift in automation and decision-making. These AI agents combine language understanding with tool use and planning capabilities, enabling them to operate in dynamic, open-world settings with minimal human supervision \cite{hughes2025ai, reed2022generalist}. Frameworks like ReAct, Chain-of-Thought (CoT), and STaR have demonstrated LLMs’ ability to reason, explore decision spaces, and adapt to novel tasks by generating interpretable and actionable outputs \cite{yao2023react, wei2022chain, zelikman2022star}. However, their application in performance-critical domains such as model compression remains underexplored, particularly in workflows that require structured, machine-executable outputs rather than natural language responses. In this work, we develop a profiling-guided agentic compression approach that automates the end-to-end optimization of vision models. Our contributions are:

\begin{enumerate}[leftmargin=*]
    \item An automated pipeline that profiles CNN and ViT models using static (MACs, parameters) and dynamic (latency, memory) metrics to identify performance-critical layers for targeted compression.
    \item 
    A novel agentic approach that analyzes profiling logs to recommend compression strategies tailored to specific models for structured layer-wise pruning and post-training dynamic quantization.
    \item An Auto Pruning Agent that performs multi-round, feedback-guided pruning to adaptively discover pruned models that optimally balance accuracy and latency.
\end{enumerate}

\section{Related Works}
Compression techniques such as pruning and quantization are widely used to reduce the computational and memory footprint of deep neural networks. Structured pruning removes large network components like filters or layers to maintain architectural regularity \cite{cheng2024survey}. Recent methods like ShortGPT \cite{men2024shortgpt} removes redundant layers using learned importance scores, and LLM-Pruner \cite{ma2023llm} prunes coupled structures using gradient-based sensitivity, SuperSAM\cite{abebe2025supersam} converts the Segment Anything Model (SAM) into a weight-sharing supernetwork using structured layer pruning and parameter prioritization. L1-norm pruning removes weights with the smallest absolute values \cite{li2017pruning}, and L2-norm pruning uses the Euclidean norm to achieve smoother sparsity and better stability in CNNs \cite{he2019filter}. Random pruning, by contrast, removes weights or filters uniformly without regard to importance, serving as a lower-bound baseline \cite{gale2019state}.
Quantization reduces model precision to lower memory and compute costs. It is commonly implemented via PTQ \cite{nagel2020up, li2021brecq, wei2022qdrop}, which calibrates quantized weights using representative data, or QAT \cite{choi2018pact, bhalgat2020lsq+, jung2019learning, esser2019learned, shin2023nipq}, which inserts quantization operators during training to adapt model parameters via gradient updates. Beyond static quantization, dynamic quantization adjusts quantization intervals based on input-dependent activations to reduce error \cite{zhong2022dynamic, lee2023insta, so2023temporal}. MergeQuant \cite{wang2025mergequant} integrates per-channel static quantization with linear mappings to reduce overhead. Our framework builds on this by applying PTQ with full dynamic quantization across all linear layers.

LLMs are increasingly being used not only as generative tools, but as autonomous agents capable of making decisions in complex, goal-driven environments. Notable prompting strategies such as CoT \cite{wei2022chain, madaan2022text} demonstrates that LLMs can create their own thinking procedure for solving problems. ReAct \cite{yao2023react} and STaR \cite{zelikman2022star} show that LLMs can self-generate intermediate reasoning steps to improve interpretability and accuracy in problem-solving. These capabilities have been extended to settings that require sequential planning, tool use, and API invocation, enabling LLMs to interact with and manipulate external systems \cite{ahn2022can, lin2023swiftsage, park2023generative, schick2023toolformer}. Unlike traditional rule-based systems, these agents adapt to new tasks by leveraging in-context learning and do not require retraining. Our work extends this paradigm to the domain of machine learning system optimization, where the agent reasons over structured performance traces and architectural metadata to generate actionable strategies for pruning and quantization.

\section{Motivation}
\begin{figure}[t]
  \centering
  \includegraphics[width=\linewidth]{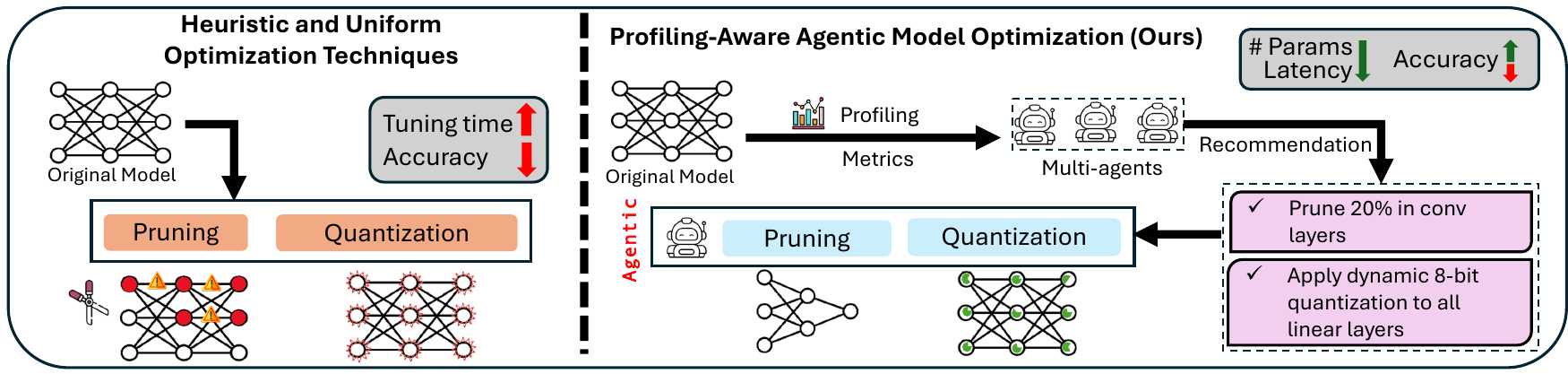}
  \caption{Heuristic vs. agentic optimization: our method uses profiling-guided optimization.}
  \label{fig:framework_overview1}
\end{figure}
While pruning and quantization are widely used for compressing models in resource-constrained settings, most approaches rely on \textit{heuristic, greedy, or trial-and-error} searches such as L1-norm pruning which remove low-magnitude weights/channels under the assumption that they contribute least to model performance. While efficient, these strategies are input-agnostic, overlook inter-layer dependency, and ignore runtime bottlenecks like latency or memory. Moreover, these methods are computationally expensive, requiring iterative retraining, grid searches, or exhaustive calibration which consumes significant GPU time and limits scalability. Compression is typically applied uniformly or per-layer without accounting for architectural inefficiencies or hardware constraints. We address these limitations by introducing a modular agentic approach (Figure~\ref{fig:framework_overview1}) that performs targeted, profiling-guided  optimization. Instead of relying on weight magnitudes or exhaustive search, we begin by profiling runtime behavior to collect hardware-relevant metrics such as MACs, latency, and memory usage. These metrics are then passed to LLM-based agents, which infer pruning and quantization strategies based on observed bottlenecks. This allows the compression to adapt to model architecture and deployment constraints by using performance profiling as a primary signal in compression. Tools like PyTorch Profiler \cite{pytorch_profiler}, TensorFlow Profiler \cite{tensorflow_profiler}, NVIDIA NVProf \cite{nvidia_profiler}, and CUPTI \cite{nvidia_cupti} expose fine-grained runtime traces including CPU/GPU utilization, memory usage, and operator-level latency but are typically used for manual inspection rather than automated compression \cite{hu2022dpro}. We directly incorporate this low-level feedback into the compression loop, enabling strategies that prioritize runtime-critical layers, unlike uniform approaches such as L1-norm pruning. Designing such a framework introduces two key challenges: (1) generalizing across architectures (CNNs and ViTs) requires unified interfaces for profiling, compression, and evaluation; and (2) integrating LLMs as decision agents demands robust prompt and parser design, as language models are not inherently optimized for structured, machine-readable outputs.

\section{Agentic Model Optimization}
\begin{figure}[t]
  \centering
  \includegraphics[width=\linewidth]{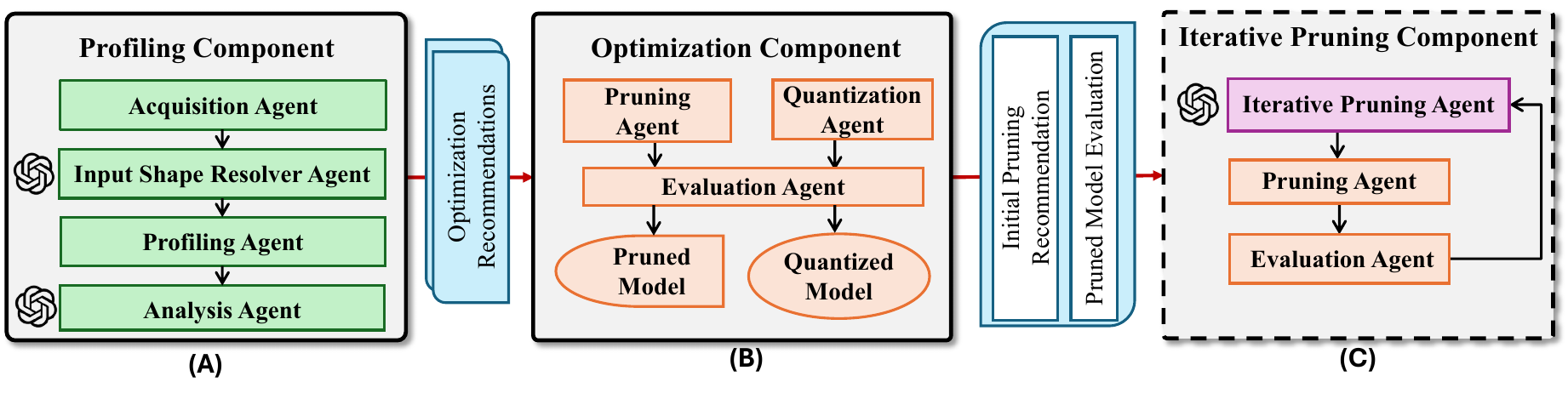}
  \caption{Overview of the ProfilingAgent,  profiling, optimization, and iterative pruning components. }
  \label{fig:framework_overview}
\end{figure}

We propose a modular, agentic approach for automated profiling-aware optimization, termed \textbf{ProfilingAgent}, designed for vision classification models, including both CNNs and ViTs. Our approach decomposes the optimization process into a sequence of discrete sub-tasks, with each agent independently responsible for a specific sub-task. \textbf{ProfilingAgent}(See Figure~\ref{fig:framework_overview}) consists of three principal components: (1) \textbf{Profiling}, which loads models, applies profiling tools to identify performance bottlenecks, and conducts analyses to generate optimization recommendations; (2) \textbf{Optimization}, which applies model compression techniques such as pruning and quantization based on the recommendations provided by the profiling component; and (3) \textbf{Iterative Pruning}, an optional component that searches for the optimal pruned model balancing accuracy and latency. Users may choose to apply only the profiling and optimization stages, or optionally engage the iterative pruning stage for further refinement.

\begin{figure}[h]
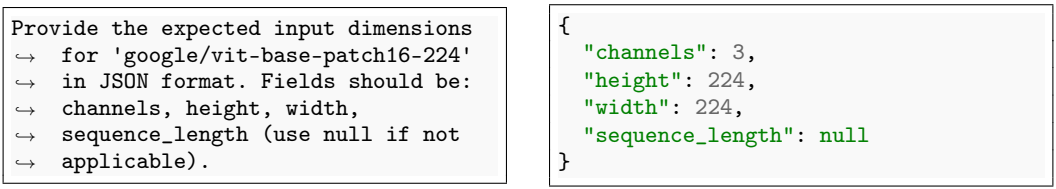

\centering
\begin{minipage}{0.48\linewidth}
\begin{minted}[
  fontsize=\small,
  frame=single,
  breaklines=true,
  breakanywhere=true,
  bgcolor=gray!5
]{text}
Provide the expected input dimensions for 'google/vit-base-patch16-224' in JSON format. Fields should be: channels, height, width, sequence_length (use null if not applicable).
\end{minted}
\end{minipage}
\hfill
\begin{minipage}{0.48\linewidth}
\begin{minted}[
  fontsize=\small,
  frame=single,
  breaklines=true,
  breakanywhere=true,
  bgcolor=gray!5
]{json}
{
  "channels": 3,
  "height": 224,
  "width": 224,
  "sequence_length": null
}
\end{minted}
\end{minipage}
\caption{Prompt issued to the LLM (left) and its structured JSON response (right).}\vspace{-1.2em}
\label{fig:llm_prompt_response}
\end{figure}

\subsection{Profiling Component}
This component (Figure~\ref{fig:framework_overview}A) includes four agents responsible for model setup and profiling. The \textbf{Acquisition Agent} initiates the pipeline by retrieving a pretrained model and its processor from Hugging Face\cite{huggingface_transformers}. It (1) configures the image processor to match the model’s preprocessing needs and (2) moves both model and processor to the target device (CPU or GPU), ensuring compatibility for downstream agents.
The \textbf{Input Shape Resolver Agent} automates identifying input tensor dimension for vision models, a prerequisite for accurate profiling that otherwise requires manual inspection. Classified as an \textit{LLM-Interactive Agent}, it uses GPT-4o to generate valid input configurations from minimal context. Based on architectural metadata, the agent constructs a prompt, submits it to the LLM, and receives input shape fields (e.g., height, width, channels) in structured JSON format (Figure.~\ref{fig:llm_prompt_response}). The result is stored for downstream use, enabling model-compatible profiling without human intervention.
The \textbf{ProfilingAgent} characterizes model complexity through both static and dynamic profiling. For theoretical estimates, it uses ptflops \cite{ptflops} to compute \textit{multiply-accumulate operations (MACs)}, an approximation of the total floating-point operations required for a single forward pass, and \textit{parameter counts}, the total number of trainable weights in the model. For empirical metrics, it invokes the PyTorch Profiler \cite{pytorch_profiler}on both CPU and GPU to capture operation-level latency, memory usage, tensor shapes, and call stacks during inference. The agent also performs manual per-layer profiling across submodules. All results are serialized and stored for downstream use by the \textbf{Analysis Agent}, enabling reproducible comparisons across models or compression states. The \textit{Analysis Agent} interprets the profiling report generated by the previous stage and synthesizes a model-specific optimization plan. 
This agent, categorized as an \textit{LLM-Interactive Agent}, acts as the core of the system by querying a LLM to propose optimization strategies based on both architectural and runtime characteristics. 
Given this report, the agent constructs a detailed prompt (see Figure~\ref{fig:prompt_example_resnet}) and instructs the LLM to return optimization suggestions in a structured JSON format. 
Pruning suggestions specify candidate layers, pruning types (structured or head pruning), pruning ratios, and brief justifications. 
Quantization suggestions are restricted to dynamic quantization methods, specifying the target layers, data types (either \texttt{qint8} or \texttt{float16}), and accompanying rationales.
These outputs form the foundation for applying downstream compression techniques.

\begin{figure}[t]
  \centering
  {\small                          
    \begin{minipage}{\linewidth}
     \begin{minted}[
          fontsize=\small,
          frame=single,
          breaklines=true,
          breakanywhere=false,
          bgcolor=gray!5
]{json}

  "pruning_recommendations": 
      {"layer": "encoder\\.stages\\d+\\.layer\\.\\d+\\.convolution",
      "pruning_type": "structured",
      "pruning_ratio": 0.2,
      "justification": "Reduces Conv2d parameters and MACs, improving latency"}
  "quantization_recommendations": 
      {"layer": "all",
      "quantization_type": "dynamic",
      "dtype": "qint8",
      "justification": "qint8 quantization of linear layers improves CPU inference"}
      \end{minted}
    \end{minipage}
  }
  \caption{Example output from the Analysis Agent for a ResNet model. The LLM recommends structured pruning and dynamic quantization strategies.}
  \label{fig:prompt_example_resnet}

\end{figure}


\subsection{Optimization Component}
The Optimization Component (Figure~\ref{fig:framework_overview}B) consists of three agents. The \textbf{Pruning Agent} applies structured or head pruning to pretrained vision models based on model-specific optimization recommendations from the \textit{Analysis Agent}. It loads the original model from Hugging Face and creates an intact copy for baseline evaluation. Pruning decisions take profiling into account and utilize previous analyses of model complexity and bottlenecks to inform the choice of pruning targets.  The agent then constructs a \texttt{DependencyGraph} to perform safe, dependency-aware pruning across interconnected layers. Pruning instructions specified in structured JSON, include layer patterns, pruning types (structured or head), and pruning ratios. Each of these recommendations details a layer pattern, a pruning type (either structured or head pruning), and a target pruning ratio represented as a float value between 0 and 1. Structured pruning removes output channels from convolutional and linear layers, while head pruning eliminates attention heads. Layer patterns are matched dynamically using regular expressions. After applying the pruning, the agent saves the optimized model for downstream evaluation. The \textbf{Quantization Agent} applies dynamic quantization to pretrained vision models using profiling-informed recommendations from the \textit{Analysis Agent}. Upon initialization, it loads the model from Hugging Face and identifies all eligible \texttt{nn.Linear} layers. Quantization policies, parsed from a structured JSON block, specify either full-model or layerwise quantization, the target datatype (\texttt{qint8} or \texttt{float16}), and brief justifications. Full-model quantization applies to all linear layers, while selective quantization uses regex-matched layer names. The agent compresses weights and accelerates inference via PyTorch’s \texttt{quantize\_dynamic}, particularly for CPU-bound scenarios. If no recommendations are provided, it optionally falls back to user-defined command-line arguments. The quantized model is saved for evaluation. The \textbf{Evaluation Agent} benchmarks the original and optimized models on a user-specified image classification dataset. It loads data from Hugging Face, selects a random subset for fair comparison and reduce overhead, and moves each model to evaluation mode. Inputs are preprocessed using the appropriate \texttt{AutoImageProcessor}, followed by forward passes to record predicted labels and inference times using high-resolution timers. Accuracy is computed via substring matching to account for minor label formatting differences. The agent also measures memory usage by summing the sizes of all parameters and buffers, and reports total parameter count as a proxy for model complexity. A detailed report is generated comparing latency, accuracy, memory consumption (in MB), and parameter count across model variants.

\begin{algorithm}[t]
\caption{Iterative Structured Pruning Agent Algorithm \\
\textbf{Input:} Profiling data $P$, Initial pruned model $M_0$, Eval. Agent, Pruning Agent, max iterations $T$ \\
\textbf{Output:} Best pruned model $M_{best}$, pruning config $R_{best}$, eval. logs $E_{best}$}
\begin{algorithmic}[1]
    \State Evaluate $M_0$: $(lat_0, acc_0) \gets \text{Evaluate}(M_0)$
    \State $M_{best} \gets M_0$, $acc_{best} \gets acc_0$, $params_{best} \gets \text{Params}(M_0)$, $lat_{best} \gets lat_0$
    \For{$i = 1$ to $T$}
        \If{$i == 1$} \State Use baseline analysis logs \EndIf
        \State Load previous analysis $A_{i-1}$ and evaluation $E_{i-1}$
        \State Query LLM with $(P, A_{i-1}, E_{i-1})$ to get $R_i$
        \State Apply $R_i$ to $M_0$ via \textbf{PruneAgent} to obtain $M_i$
        \State Evaluate $M_i$: $(lat_i, acc_i) \gets \text{Evaluate}(M_i)$, $params_i \gets \text{Params}(M_i)$
        \If{Better($acc_i, params_i, lat_i$)}
            \State $M_{best}, acc_{best}, params_{best}, lat_{best}, R_{best} \gets \text{Update If Better}$
        \EndIf
    \EndFor
    \State Save and \Return $M_{best}, R_{best}, E_{best}$

\end{algorithmic}
\label{alg:iterative}
\end{algorithm}

\subsection{Iterative Pruning Component}
At the end of the optimization component, the user has the option to continue by activating the \textit{Iterative Pruning Component} (Figure~\ref{fig:framework_overview}C), which consists of one agent: the \textbf{Iterative Pruning Agent}, managing model optimization across multiple rounds. By engaging with the \textit{Pruning Agent} and \textit{Evaluation Agent} at each iteration, the Iterative Pruning Agent automates the processes of pruning, evaluating, and generating recommendations, thereby refining the model progressively and determining the best-pruned version. 
By supervising the interactions among profiling data, the \textit{Pruning Agent}, and the \textit{Evaluation Agent}; the \textit{Iterative Pruning Agent} automates the processes of multi-round pruning and evaluation. 
The agent begins with an initial profiling report and baseline pruning evaluation (\texttt{Line} 5 in Algorithm \ref{alg:iterative}), then iteratively queries a LLM to produce updated pruning recommendations derived in the model's prior pruning configurations, evaluation results, and static profiling features (\texttt{Line} 3-11). 
During every iteration, the agent implements the updated pruning plan via the \textit{Pruning Agent} (\texttt{Line} 8), reassesses the pruned model with the \textit{Evaluation Agent} (\texttt{Line} 9), and retrieves essential performance metrics like inference latency, classification accuracy, and parameter count.
The agent keeps track of the best model discovered over iterations (\texttt{Line} 10-11) through a hierarchical selection strategy that prioritizes higher accuracy first, then lower parameter count, and finally faster inference time.
Once the required number of iterations is finished, the agent saves the best-pruned model along with its corresponding pruning analysis file and evaluation report (\texttt{Line} 12). 
 This optimization framework, which operates on a closed loop, allows for the dynamic adjustment of pruning strategies and enhances model efficiency using empirical performance feedback.

\section{Results}
\subsection{Experimental Setup}
\noindent \textbf{Datasets and Models. }We evaluate our framework on three standard image classification datasets loaded from Hugging Face \cite{polhugging}, Imagenette \cite{Howard_Imagenette_2019}, CIFAR-10, and CIFAR-100 \cite{krizhevsky2009learning}. Imagenette is a simplified 10-class subset of ImageNet \cite{deng2009imagenet} with 13k high-resolution images (320×320). CIFAR-10 consists of 60k 32×32 images across 10 categories, while CIFAR-100 offers finer granularity with 100 classes and 600 images each. In addition, we also evaluate directly on the full ImageNet-1K dataset \cite{deng2009imagenet} , which contains 1.2 million labeled images across 1k classes. We benchmark four pretrained models on ImageNet-1K  ResNet-101, ViT-B/16, Swin-base, and DeiT-B/16.
ResNet-101 \cite{he2016deep, aksoy2022prediction} is a 101-layer convolutional network based on ResNet-50 \cite{ghosal2019brain} with residual blocks that improve training stability and performance. ViT-B/16 \cite{barulina2022deep, morani2023covid} tokenizes image patches and processes them via transformer encoders. Swin-base \cite{liu2021swin} extends ViT with hierarchical feature maps and window based attention for better efficiency and multi scale representation. DeiT-B/16 \cite{touvron2021training} follows the standard ViT design but is trained more efficiently using knowledge distillation.

\noindent \textbf{Hardware/Platform. }The NCSA Delta supercomputing system was used to conduct all experiments. The primary profiling and optimization tasks were carried out on compute nodes with AMD EPYC 7763 64-core CPUs, NVIDIA A100-SXM4-40GB GPUs, and 251 GB of system RAM. The fine-tuning experiments on CIFAR-10 and CIFAR-100 were conducted on Delta nodes equipped with NVIDIA H200 GPUs (144 GB memory) and dual Intel Xeon Platinum 8558 CPUs (totaling 96 cores), backed by 2 TB of RAM. The software environment featured CUDA v11.8 and PyTorch v2.6.0, utilizing GPU acceleration throughout all training and evaluation phases.

\noindent \textbf{Baselines. }
We implemented three structured pruning baselines, L1, L2 \cite{li2016pruning}, and random, that remove entire filters or heads, preserving architecture and enabling speedups without sparse libraries \cite{cheng2024survey}.
Each method eliminates output channels from convolutional or linear layers based on a predefined importance metric. In L1-norm pruning, importance is computed by summing the absolute weights per output channel. For convolutional layers in ResNet, this corresponds to each output filter; for MLP layers in Swin, ViT, and DeiT, it involves summing across weight matrix rows. Channels with the lowest L1 scores are pruned under the assumption that smaller absolute weights contribute less to the model's output. L2-norm pruning instead computes the Euclidean norm, yielding smoother sparsity and improved stability, particularly in convolutional layers \cite{mehta2019implicit}. While the ranking criterion differs, its pruning logic mirrors the L1 approach. Random pruning, by contrast, removes output channels uniformly at random, ignoring magnitude or structure, and serves as a lower-bound baseline to evaluate the benefits of magnitude-aware pruning. All three methods use dependency graphs to ensure structurally consistent pruning throughout the model architecture.
To evaluate our quantization approach, we use ONNX Runtime’s \cite{onnxruntime2023quantization}  PTQ as our baseline. It dynamically quantizes pretrained models by converting \texttt{MatMul} and \texttt{Gemm} weights to \texttt{8-bit} integers (\texttt{QInt8}) without the need for retraining.  While \texttt{MatMul} performs standard matrix multiplication, \texttt{Gemm} extends it with optional scaling and bias, which are standard in fully connected layers.

\subsection{Pruning Evaluation On Imagenette}

\begin{table}[H]
  \caption{Impact of pruning on $\Delta$Acc. and parameter counts.}
  \label{tab:pruning-Image10}
  \centering
  \small
  \begin{tabular}{%
    l  
    l  
    *{4}{cc} 
  }
    \toprule
    \multirow{2}{*}{Type} 
      & \multirow{2}{*}{Ratio} 
      & \multicolumn{2}{c}{DeiT-B/16 (86.6M)} 
      & \multicolumn{2}{c}{ResNet-101 (44.5M)} 
      & \multicolumn{2}{c}{Swin-B (87.8M)} 
      & \multicolumn{2}{c}{ViT-B/16 (86.6M)} \\
    \cmidrule(lr){3-4} \cmidrule(lr){5-6} \cmidrule(lr){7-8} \cmidrule(lr){9-10}
      & 
      & $\Delta$Acc. & \#P (M)
      & $\Delta$Acc. & \#P (M)
      & $\Delta$Acc. & \#P (M)
      & $\Delta$Acc. & \#P (M) \\
    \midrule
    \multirow{3}{*}{L1 Norm}
      & 1\%   & -1.0  & 86.0  & -13.0 & 40.6  & -1.0  & 87.2  &  0.0  & 86.0  \\
      & 10\%  & -3.0  & 80.9  & -17.0 & 23.4  &  0.0  & 82.2  &  0.0  & 80.9  \\
      & 20\%  & -7.0  & 75.2  & -17.0 & 16.6  & -3.0  & 76.6  & -3.0  & 75.2  \\
    \midrule
    \multirow{3}{*}{L2 Norm}
      & 1\%   &  0.0  & 86.0  & -17.0 & 40.6  & -1.0  & 87.2  &  0.0  & 86.0  \\
      & 10\%  & -1.0  & 80.9  & -17.0 & 23.4  & -2.0  & 82.2  &  0.0  & 80.9  \\
      & 20\%  & -9.0  & 75.2  & -17.0 & 16.6  & -4.0  & 76.6  & -3.0  & 75.2  \\
    \midrule
    \multirow{3}{*}{Random}
      & 1\%   &  0.0  & 86.0  & -11.0 & 40.6  &  0.0  & 87.2  &  0.0  & 86.0  \\
      & 10\%  & -3.0  & 80.9  & -17.0 & 23.4  &  1.0  & 82.2  &  0.0  & 80.9  \\
      & 20\%  & -3.0  & 75.2  & -17.0 & 16.6  & -3.0  & 76.6  & -2.0  & 75.2  \\
    \midrule
    Ours 
      & —     & -1.0  & 80.9  & +1.0  & 43.0  & -1.0  & 81.1  & +2.0  & 78.1  \\
    \bottomrule
  \end{tabular}
\end{table}

\subsubsection{Accuracy and Parameter Reduction}
Our pruning agent’s test accuracy after five iterations, as well as that of three baseline methods (L1-norm, L2-norm, and random pruning), is shown in Table~\ref{tab:pruning-Image10}.  Our agentic framework adaptively modifies the pruning ratio and the set of prunable layers at every iteration, based on profiling insights.  After five iterations, the agent arrives at an optimized pruning strategy that is customized to fit the model's structure and performance characteristics.  Conversely, baseline methods use a uniform fixed pruning ratio for all layers that can be pruned.  We manually applied baseline pruning ratios at levels of 20\%, 10\%, and 1\% to ensure a fair comparison, and reported the performance of the resulting models across these configurations.

Table~\ref{tab:pruning-Image10} illustrates the impact of our iterative, agentic pruning strategy in comparison to L1-norm, L2-norm, and random pruning methods across four distinct models. In contrast to baseline techniques that implement fixed pruning ratios uniformly across layers, results in nearly identical parameter reductions, as they eliminate the same proportion of output channels irrespective of layer significance but our strategy adaptively determines pruning configurations based on iterative profiling and performance feedback. This approach facilitates consistent retention-or even enhancement-of top-1 accuracy, while achieving comparable or superior parameter reduction. For example, our method maintains full model functionality in ResNet-101 (43M parameters, +1\% $\Delta$ accuracy), whereas norm-based and random strategies often lead to network collapse under elevated pruning ratios. Furthermore, although random pruning occasionally exceeds norm-based pruning in accuracy (e.g., Swin-B at 10\% pruning ratio), our method consistently attains a more advantageous balance between parameter reduction and accuracy across all models, thereby underscoring the advantages of profiling-aware and dynamically guided pruning.

\subsubsection{Inference Speedup and Memory Usage Reduction}
\begin{figure}[t]
  \centering
  \includegraphics[width=\textwidth]{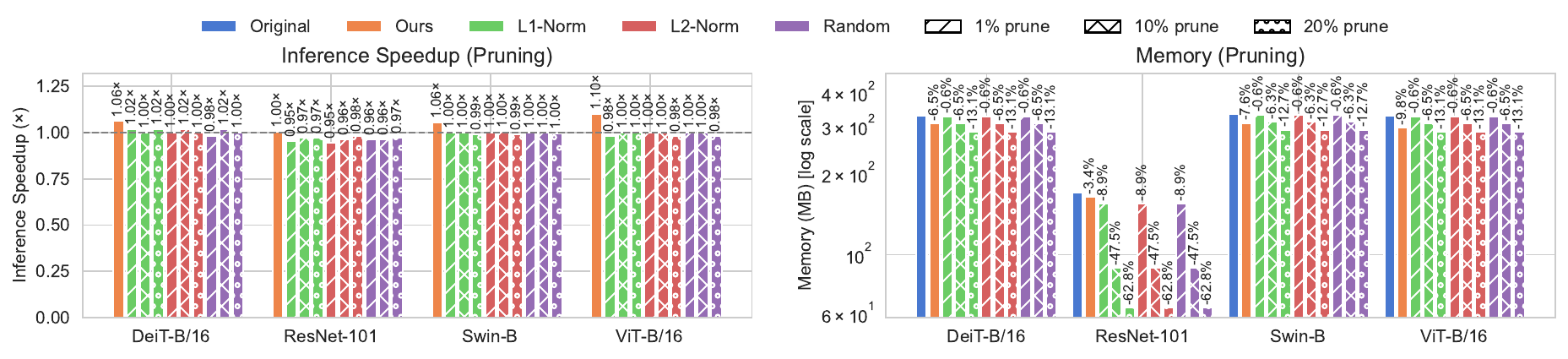}
  \caption{Inference speedup and memory reduction of our pruning vs. L1, L2, and random baselines.}
  \label{fig:imagenet}
\end{figure}

Figure~\ref{fig:imagenet} illustrates that our pruning technique offers greater inference speedups (up to $1.10\times$) compared to all baselines, while still ensuring competitive memory reductions. The slight slowdown observed in ResNet-101 results from misaligned channels and additional overhead. In summary, our method achieves a more effective balance between compression and performance through adaptive, profiling-guided  pruning.

\subsection{Generalization of Our Pruning Method to CIFAR-10/100}

To further assess the generalizability of our pruning approach beyond large-scale benchmarks, we evaluate on CIFAR-10 and CIFAR-100 without applying any fine-tuning after pruning. As shown in Table~\ref{tab:cifar10-100-pruning-summary}, our method consistently reduces both parameter counts and model storage while maintaining competitive accuracy across diverse architectures. For instance, Swin-Base achieves a 7.6–7.7\% memory reduction with less than a 1\% accuracy drop on both datasets, and ResNet-101 even shows a slight accuracy improvement on CIFAR-10. These results demonstrate that our pruning strategy is not limited to ImageNet-1K but transfers effectively to smaller-scale datasets, underscoring its robustness and applicability across a wide range of vision tasks.

\begin{table}[H]
\centering
\small
\caption{Pruning results on CIFAR-10 and CIFAR-100. Top-1 shows accuracy before $\rightarrow$ after pruning. Parameters are reported in millions (M). Memory reduction is based on model file size (MB).}
\label{tab:cifar10-100-pruning-summary}
\begin{tabular}{@{} l l c c c @{}}
\toprule
Model & Dataset & Top-1 Acc.\, & Params (M) & Mem.\,Red.\,$\downarrow$\,(\%) \\
\midrule
\midrule
ResNet-101 & CIFAR-10  & 84.10 $\rightarrow$ 84.20 & 48.81 $\rightarrow$ 47.30 & 3.1 \\
           & CIFAR-100 & 83.30 $\rightarrow$ 78.40 & 42.71 $\rightarrow$ 41.20 & 3.5 \\
\midrule
ViT-B/16   & CIFAR-10  & 98.50 $\rightarrow$ 97.90 & 87.12 $\rightarrow$ 81.45 & 6.5 \\
           & CIFAR-100 & 90.00 $\rightarrow$ 88.10 & 85.88 $\rightarrow$ 83.04 & 3.3 \\
\midrule
Swin-Base  & CIFAR-10  & 98.60 $\rightarrow$ 97.90 & 88.32 $\rightarrow$ 81.61 & 7.6 \\
           & CIFAR-100 & 87.70 $\rightarrow$ 87.00 & 86.85 $\rightarrow$ 80.13 & 7.7 \\
\midrule
DeiT-B/16  & CIFAR-10  & 97.80 $\rightarrow$ 97.10 & 87.12 $\rightarrow$ 81.46 & 6.5 \\
           & CIFAR-100 & 85.50 $\rightarrow$ 84.90 & 85.95 $\rightarrow$ 84.26 & 2.0 \\
\bottomrule
\end{tabular}
\end{table}

\subsection{Quantization Evaluation}
Table~\ref{tab:quant-summary} presents a comparison of our quantization agent with the dynamic quantization of ONNX Runtime, utilizing pretrained ImageNet-1K models on the ImageNet-1K dataset without undergoing fine-tuning. Both methodologies implement dynamic PTQ with QInt8 precision. Our agent utilizes profiling-guided analysis to carry out full dynamic quantization across all torch.nn.Linear layers, as consistently recommended by the analysis agent based on profiling logs. Across all ImageNet-1K models, our method preserves accuracy relative to ONNX, with accuracy drops of less than 0.5\% in nearly all cases (reported as rounded values, yielding at most a 1\% drop for DeiT-B/16).At the same time, it consistently achieves greater memory reductions, e.g., 74.7\% vs. 73.0\% for Swin-Base, along with substantial inference speed gains ranging from 1.36× (ResNet-101) to 1.66× (ViT-B/16).These findings underscore the advantages of profiling-guided , layer-selective quantization in comparison to uniform ONNX strategies.

\begin{table}[H]
\centering
\small
\caption{Post-training quantization results on pretrained ImageNet-1K models. Metrics include $\Delta$Acc., memory reduction, and inference latency. “Speed Gain vs ONNX” denotes the ratio of ONNX Runtime’s average inference time to ours.}
\label{tab:quant-summary}
\begin{tabular}{@{} l l c c c c @{}}
\toprule
Model & Method 
  & $\Delta$Acc.\,(\%) 
  & Mem.\,Red.\,$\downarrow$\,(\%) 
  & Avg.\,Inference (s) 
  & Speed Gain vs ONNX \\
\midrule
\midrule
ViT-B/16 & ONNX   &  0  & 74.1 & 0.3869 $\rightarrow$ 0.2179 & -- \\
         & Ours   &  0  & 74.2 & 0.2320 $\rightarrow$ 0.1316 & 1.66$\times$ \\
\midrule
ResNet-101 & ONNX   &  0  & 3.4  & 0.1640 $\rightarrow$ 0.1634 & -- \\
           & Ours   &  0  & 3.4  & 0.1211 $\rightarrow$ 0.1199 & 1.36$\times$ \\
\midrule
Swin-Base & ONNX   &  0  & 73.0 & 0.3749 $\rightarrow$ 0.2266 & -- \\
          & Ours   &  0  & 74.7 & 0.2442 $\rightarrow$ 0.1556 & 1.46$\times$ \\
\midrule
DeiT-B/16 & ONNX   & 0  & 74.1 & 0.3863 $\rightarrow$ 0.2174 & -- \\
          & Ours   & -1  & 74.2 & 0.2312 $\rightarrow$ 0.1317 & 1.65$\times$ \\
\bottomrule
\end{tabular}
\end{table}


\subsection{Pruning and Quantization Evaluation on ImageNet-1K}

We further evaluate the effect of our pruning and quantization strategies on the ImageNet-1K dataset across multiple architectures, including ViT-B/16, ResNet-101, Swin-Base, and DeiT-B/16. As summarized in Table~\ref{tab:imagenet-quant-prune}, pruning achieves moderate reductions in parameters and memory (e.g., 3.3\% for ViT-B/16 and 2.9\% for ResNet-101) with only minor accuracy degradation (less than 1\% in most cases). In contrast, quantization yields substantially larger memory savings (approximately 74\% reduction across models) and consistent inference speedups, reaching up to 1.74× on ViT-B/16 and 1.73× on DeiT-B/16, while largely preserving Top-1 accuracy. These results demonstrate the complementary strengths of pruning and quantization: pruning reduces model size with minimal accuracy loss, whereas quantization delivers significant efficiency gains in memory and latency, underscoring the effectiveness of our method for scaling ImageNet-1K models to resource-constrained environments.

\begin{table}[H]
\centering
\small
\caption{Effect of pruning and quantization on Top-1 accuracy, memory, parameter count, and inference latency on ImageNet-1K.}
\label{tab:imagenet-quant-prune}
\begin{tabular}{@{} l l c c c c @{}}
\toprule
Model & Method 
  & Top-1 Acc.\,(\%) 
  & Mem.\,Red.\,$\downarrow$\,(\%) 
  & Param.\,Red.\,$\downarrow$\,(\%) 
  & Speed-up \\
\midrule
\midrule
ViT-B/16 & Pruning   & 80.3 $\rightarrow$ 79.5 &   3.3 &   3.3 & 1.02$\times$ \\
         & Quant.   & 80.3 $\rightarrow$ 80.2 &  74.3 &   0.0 & 1.74$\times$ \\
\midrule
ResNet-101 & Pruning & 82.4 $\rightarrow$ 79.7 &   2.9 &   2.9 & 0.97$\times$ \\
           & Quant. & 82.4 $\rightarrow$ 82.2 &   3.4 &   0.0 & 1.01$\times$ \\
\midrule
Swin-Base & Pruning  & 85.3 $\rightarrow$ 84.7 &   1.9 &   1.9 & 1.02$\times$ \\
          & Quant.  & 85.3 $\rightarrow$ 85.1 &  74.7 &   0.0 & 1.59$\times$ \\
\midrule
DeiT-B/16 & Pruning  & 81.0 $\rightarrow$ 80.2 &   2.0 &   2.0 & 0.98$\times$ \\
          & Quant.  & 81.0 $\rightarrow$ 79.8 &  74.3 &   0.0 & 1.73$\times$ \\
\bottomrule
\end{tabular}
\end{table}

\subsection{GPT-4o vs GPT-4-Turbo}

Our Iterative Pruning Agent integrates an LLM within a closed-loop system alongside pruning and evaluation modules. This system utilizes profiling data and prior decisions to enhance pruning strategies. In order to evaluate the effectiveness of LLM reasoning, we conducted a comparative analysis of GPT-4o and GPT-4-Turbo. The results indicated that GPT-4o achieved more stable and contextually aware pruning, thereby maintaining accuracy (for example, ViT-B/16: +2\%). In contrast, GPT-4-Turbo frequently produced excessively aggressive plans, which significantly degraded performance (for instance, ResNet-101: -14\%). A summary of the results is presented in Table~\ref{tab:gpt-platform-comparison}.

\begin{table}[H]
\centering
\small
\setlength{\tabcolsep}{5pt} 
\caption{Comparison of pruning outcomes: GPT-4o vs. GPT-4-Turbo.}
\label{tab:gpt-platform-comparison}
\begin{tabular}{l *{2}{ccc}}
\toprule
Model 
  & \multicolumn{3}{c}{\textbf{GPT-4o}} 
  & \multicolumn{3}{c}{\textbf{GPT-4-Turbo}} \\
\cmidrule(lr){2-4} \cmidrule(lr){5-7}
  & $\Delta$ Acc.\,(\%) 
  & Orig.\ \#P(M) 
  & \#P(M) $\downarrow$ 
  & $\Delta$ Acc.\,(\%) 
  & Orig.\ \#P(M) 
  & \#P(M) $\downarrow$ \\
\midrule
DeiT-B/16      
  & $-1$  & $86.56$ & $80.9$  
  & $-4$  & $86.56$ & $78.06$  \\
ResNet-101     
  & $+1$  & $44.54$ & $43.04$  
  & $-14$ & $44.54$ & $40.56$  \\
ViT-B/16  
  & $+2$  & $86.56$ & $78.06$  
  & $-1$  & $86.56$ & $80.9$  \\
Swin-B         
  & $-1$  & $87.76$ & $81.05$  
  & $-1$  & $87.76$ & $68.2$  \\

\bottomrule
\end{tabular}
\end{table}

\section{Conclusion}

We introduced ProfilingAgent, a Profiling-Guided Agentic approach for pruning and quantization to address the challenges associated with traditional optimization and harnesses the reasoning capabilities of LLMs to conduct profiling-guided optimization. This approach enables us to execute pruning and quantization with greater precision and significantly reduces time and human resources through agentic decision-making processes. We demonstrate that our approach surpasses heuristic-based optimization methods by identifying the optimal tradeoff between accuracy and latency. Furthermore, this methodology has the potential to be adapted to other optimization techniques in the future to determine the most effective optimization strategies based on model architecture.

\newpage

\end{document}